\documentclass{article}
\usepackage{arxiv}

\usepackage[utf8]{inputenc} 
\usepackage[T1]{fontenc}    
\usepackage[hidelinks]{hyperref}       
\usepackage{booktabs}       
\usepackage{amsfonts}       
\usepackage{nicefrac}       
\usepackage{microtype}      
\usepackage{lipsum}		
\usepackage{graphicx}
\usepackage[round]{natbib}
\usepackage{doi}
\usepackage{amsmath}
\usepackage{subcaption}
\usepackage{verbatim}
\usepackage{bbm}

\title{Feature Distribution Matching for Federated Domain Generalization}

\author{Yuwei Sun\\
	University of Tokyo\\
	RIKEN AIP\\
	\texttt{ywsun@g.ecc.u-tokyo.ac.jp} \\
	\And
	Ng Chong\\
	United Nations University\\
	\texttt{ngstc@unu.edu}\\
	\And
	Hideya Ochiai\\
	University of Tokyo\\
	\texttt{ochiai@elab.ic.i.u-tokyo.ac.jp}\\
}

\date{} 
\renewcommand{\headeright}{}
\renewcommand{\undertitle}{}

\begin{document}

\maketitle

\begin{abstract}
Multi-source domain adaptation has been intensively studied. The distribution shift in features inherent to specific domains causes the negative transfer problem, degrading a model’s generality to unseen tasks. In Federated Learning (FL), learned model parameters are shared to train a global model that leverages the underlying knowledge across client models trained on separate data domains. Nonetheless, the data confidentiality of FL hinders the effectiveness of traditional domain adaptation methods that require prior knowledge of different domain data. We propose a new federated domain generalization method called Federated Knowledge Alignment (FedKA). FedKA leverages feature distribution matching in a global workspace such that the global model can learn domain-invariant client features under the constraint of unknown client data. FedKA employs a federated voting mechanism that generates target domain pseudo-labels based on the consensus from clients to facilitate global model fine-tuning. We performed extensive experiments, including an ablation study, to evaluate the effectiveness of the proposed method in both image and text classification tasks using different model architectures. The empirical results show that FedKA achieves performance gains of 8.8\% and 3.5\% in Digit-Five and Office-Caltech10, respectively, and a gain of 0.7\% in Amazon Review with extremely limited training data. Moreover, we studied the effectiveness of FedKA in alleviating the negative transfer of FL based on a new criterion called Group Effect. The results show that FedKA can reduce negative transfer, improving the performance gain via model aggregation by 4 times.
\end{abstract}


\section{Introduction}
Federated learning (FL) \citep{fedavg,fl1} has been accelerating the collaboration among different institutions with a shared interest in machine learning applications such as privacy-preserving diagnosis of hospitals \citep{split} and decentralized network intrusion detection \citep{segmented}. One of the most challenging problems in FL is to improve the generality in tackling client data from different domains. These different domains are usually used for the same classification task but with particular sample features under varying data collection conditions of clients. A naive averaging of all clients' model updates cannot guarantee the global model's performance in different tasks due to the problem of negative transfer\citep{negatran}. In this regard, the learned knowledge from a client might not facilitate the learning of others. The effectiveness of model sharing in FL regarding knowledge transferability to an unseen task is of great importance to real-life application. For example, in medical diagnosis, images collected by different medical machines can vary in sample quality. Client models learned on such diverse samples can diverge in the parameter space. Simply aggregating these models will not guarantee a better global model. Another example is connected autonomous vehicles based on FL. The multi-agent systems learn to tackle different driving situations in different cities and FL allows these agents to share the experience of driving in a new city.

Feature disentanglement is a common approach to alleviating problems of domain shift and negative transfer when encountering different domains, by separating domain-invariant features and domain-specific features from training samples \citep{fd2,bkda2,bkda0,bkda1}. Nevertheless, such a practice necessitates that different domain data are centrally located at the same place for computation. In the above hospital and vehicle cases, feature disentanglement is unfeasible or impractical due to either privacy concerns or communication overheads in data sharing. The difficulty in federated domain generalization is that the source domain data of clients and the target domain data of a new task are usually separately located, which hinders effective knowledge sharing in FL. Moreover, the traditional model aggregation in approaches such as Federated Averaging (FedAvg)\citep{fedavg} cannot guarantee the improvement in the global model's performance by sharing local models trained on various client domains.

\begin{figure*}[!t]
    \centering
    \includegraphics[width=0.7\linewidth]{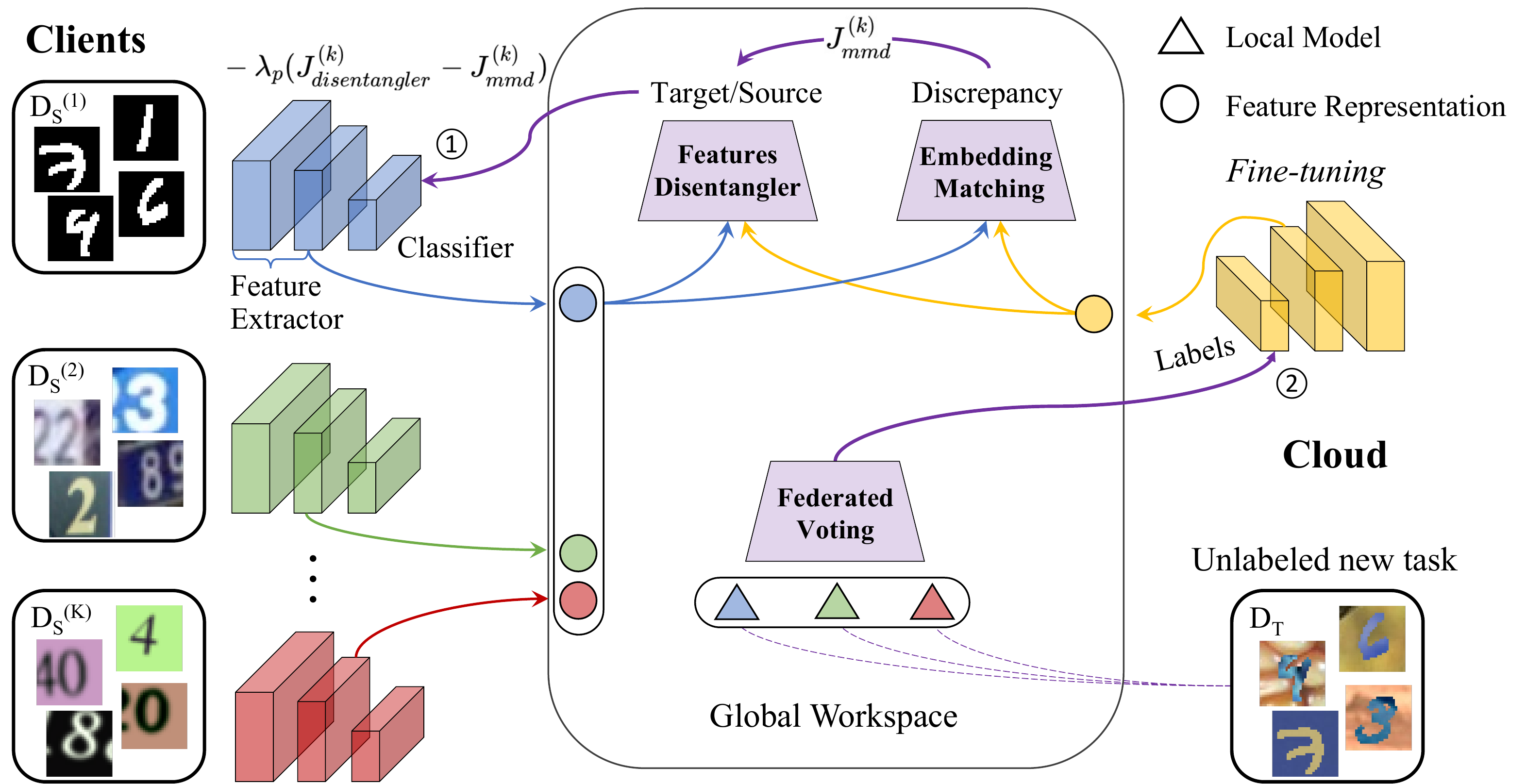}
    \caption{The framework of Federated Knowledge Alignment (FedKA) consists of three building blocks, i.e., Global Feature Disentangler, Embedding Matching, and Federated Voting. FedKA leverages distributed data domains based on feature distribution matching in a global workspace. The negative transfer is alleviated by \raisebox{.5pt}{\textcircled{\raisebox{-.9pt} {1}}} local model representation learning on client domains and \raisebox{.5pt}{\textcircled{\raisebox{-.9pt} {2}}} global model fine-tuning on the unlabeled cloud domain.}
    \label{fig:idea}
\end{figure*}

To this end, we propose Federated Knowledge Alignment (FedKA) (see Fig. \ref{fig:idea}) that alleviates negative transfer in FL improving the global model's generality to unseen tasks.

Overall, our main contributions are three-fold:

1) We proposed a novel domain generalization method FedKA in federated learning under the constraint of unknown client data, mainly due to data confidentiality. FedKA learns to reduce feature discrepancy between clients improving the global model's generality to tackle unseen tasks. (Section \ref{FedKA}). 

2) This work studied a new criterion for measuring negative transfer in federated learning (FL) called Group Effect, which throws light on the ineffectiveness of model aggregation in FL when training on different client domains. This work provided detailed formulations (Section \ref{GE}) and evaluation of Group Effect in FL (Section \ref{exp:GE}).  

3) We performed extensive experiments on three different datasets, i.e., Digit-Five, Office-Caltech10, and Amazon Review. We compared two different neural network architectures for model sharing including a lightweight two-layer model and a Resnet18 model. We demonstrated our method's effectiveness in improving the global model's prediction for various target tasks. (Section \ref{sec:exp}).

The remainder of this paper is structured as follows. Section 2 provides an overview of federated learning and domain adaptation and presents relevant work on the intersection of the two fields. Section 3 presents essential definitions and technical underpinnings of the proposed method. Section 4 presents the results of the empirical evaluation and the discussion of our main findings. Section 5 concludes the paper and provides future directions.

\section{Related Work}

\subsection{Federated Learning}
A distributed framework for machine learning (ML) \citep{dml} was introduced due to the proliferation of ML applications in academia and industry. The parameter server framework was further extended to a versatile and high-performance implementation for distributed ML based on local training data \citep{muli}. Moreover, Federated Learning (FL) \citep{fedavg} aims to train a model that learns a global probability distribution leveraging local model training on distributed data sources and trained model parameter sharing. Nevertheless, it usually bears a degraded global model performance when training on diversified client data \citep{fl2}. There have been many works studying imbalanced Non-IID data in FL \citep{noniid4,noniid3}. For instance, federated group knowledge transfer (FedGKT)\citep{He} leveraged Kullback Leibler (KL) Divergence to measure the prediction loss between an edge model and a cloud model, thus aligning knowledge of client models trained on Non-IID samples and the global model. Unfortunately, there are still not many efforts on the domain shift problem in FL, where each client owns data with domain-specific features due to different data collection environments.

\subsection{Domain Adaptation}
Domain adaptation \citep{da1,bk3,bk2} is one type of transfer learning to perform knowledge transfer from the source domain to the target domain. In this regard, a reconstruction-based method with an encoder-decoder architecture aims to learn a discriminative mapping of target samples to the source feature space, thus improving generalization performance \citep{adversarialDA,adversarialDA3}. 
However, the generative approach is usually resource-consuming relying on computational capability.
It is incompatible with resource-constrained clients in FL, such as mobile devices. In contrast, the method of feature disentanglement aims to distill features consistent across different domains thus improving the transferability of learned features. Therefore, the output of a model will remain unaffected despite several domain-specific feature changes. Deep Adaptation Networks (DAN)\citep{dan} trains two neural network models on the source and target domains, respectively. Then, DAN applies the multi-kernel Maximum Mean Discrepancy (MK-MMD) loss \citep{BorgwardtGRKSS06} to align features extracted from different layers of the two models. A variant of this method \citep{mmd2} aligns the joint distributions of multiple domain-specific layers across domains using a Joint Maximum Mean Discrepancy (JMMD) criterion. Furthermore, Domain Adversarial Neural Network (DANN)\citep{ganin} leverages the domain confusion loss and the classification loss. DANN trains a classifier that distinguishes between source domain features and target domain features with encoders that distills representations indistinguishable by the domain classifier. 

\subsection{Domain Generalization for Federated Learning}
The line of work in domain generalization for Federated Learning \citep{da1,domainshift} has been studied recently.
For example, Federated Adversarial Domain Adaptation (FADA)\citep{peng} aims to tackle domain shift in FL through adversarial learning of domain-invariant features. Moreover, \cite{yao} presented a reversed scenario of FADA, where they tackled a multi-target domain adaptation problem for transferring knowledge learned from a labeled cloud dataset to different client tasks. 

Unlike the studies mentioned above, FedKA leverages interactive learning between clients and the cloud thus overcoming the challenge of data confidentiality in FL. To improve the representation transferability, a client's encoder learns to align its output with the global embedding provided by the cloud. Simultaneously, the global model learns a better representation of the target task via fine-tuning based on the strategy of federated voting. Furthermore, to our best knowledge, there are no existing efforts to measure negative transfer in FL. Therefore, we propose Group Effect as an effective criterion.  

\section{Method}
In this section, we first define the multi-source domain adaptation problem in Federated Learning (FL). Then, we present the technical underpinnings of Federated Knowledge Alignment (FedKA). Finally, we introduce our criteria for measuring the effectiveness of domain generalization methods in FL.

\subsection{Multi-Source Domain Adaptation in Federated Learning}
We specifically consider a classification task with $C$ categories. Let $x \in \mathbb{R}^V=X$ be a sample and $y \in \{1,2,...,C\} = Y$ be a label. $D$ consists of a collection of $N$ samples as $D=\{(x_i, y_i)\}_{i=1}^N$. In unsupervised domain adaptation\citep{uda1,uda0}, given a source domain $D_S=(X_S,Y_S)$ and a target domain $D_T=(X_T,Y_T)$ where the labels $Y_T$ are not provided, the goal is to learn the target conditional probability distribution $P(Y_T|X_T)$ in $D_T$ with the information gained from $D_S$. The source domain $D_S$ and target domain $D_T$ usually share the same support of the input and output space $X\times Y$, but their data have domain discrepancies with specified styles, i.e., $P(X_S)\neq P(X_T)$. 

A federated learning (FL) framework consists of the parameter server (PS) and $K$ clients. We suppose that each client $k$ has a different source domain $D_{S}^{(k)}=(X_S^{(k)},Y_S^{(k)})$ and the PS has the unlabeled target domain $D_T=(X_T)$. Let $f$ be a neural network classifier that takes an input $x_i$ and outputs a $C$-dimensional probability vector where the $j$th element of the vector represents the probability that $x_i$ is recognized as class $j$. Then, the prediction is given by $\hat{y} = \mbox{arg}\max_j f(x)_j$ where $f(x)_j$ denotes the $j$th element of $f(x)$. 


A client usually cannot share $D_{S}^{(k)}$ with the PS nor other clients mainly due to data confidentiality. Instead, FL learns a global probability distribution by updating a global model $G_t$ based on the local models $L^{(k)}_t$ shared by different clients where $t$ denotes the time step. The model aggregation allows FL to train over the entire data without disclosing distributed training samples. Notably, FL proceeds by iterating the following steps: (1) the PS that controls the entire process of FL, initializes the global model $G_0$ and delivers it to all clients, (2) each client $k$ updates the model using $N^{(k)}$ samples from the local data $D_S^{(k)}$, and sends back its model update $L_{t+1}^{(k)}-G_t$ to the PS, (3) then, PS aggregates all local model updates based on methods such as Federated Averaging (FedAvg), updates the global model, and sends the global model to all clients. Then, the model aggregation based on FedAvg can be formulated by the following 
\begin{equation}
G_{t+1} = G_t +\sum_{k\in K}\frac{N^{(k)}}{\sum_{k\in K} N^{(k)}}(L_{t+1}^{(k)}-G_t).
\label{al:fedavg}
\end{equation}

The goal of the multi-source domain adaptation in FL is to learn a global model that predicts the target conditional probability distribution $P(Y_T|X_T)$ of $D_T$ using the knowledge from the $K$ client models learned on different source domains $\{D_{S}^{(1)},D_{S}^{(2)},...,D_{S}^{(K)}\}$. 

\subsection{Federated Knowledge Alignment}
\label{FedKA}
\subsubsection{Motivation}
Effective knowledge transfer in multi-source domain adaptation of Federated Learning (FL) is critical to the success of distributed machine learning via model sharing. The challenge is to alleviate the negative transfer of FL such that the global model's generality to unseen tasks can be improved.

We demonstrate a global workspace where different latent representations and encoder models of clients are organized in a way that they can be leveraged to perform various tasks improving the global model's generality (Figure \ref{fig:idea}). This is inspired by Global Workspace Theory\citep{Baars1988,bengio2019} which enables multiple network models to cooperate and compete in solving problems via a shared feature space for common knowledge sharing. To this end, we propose Federated Knowledge Alignment (FedKA) that leverages the representation learning of client local models and the global model by feature distribution matching in the global workspace, facilitating effective knowledge transfer between clients. 

\subsubsection{Global Feature Disentangler}
\label{sec:dis}
Let $f_{e}^{(k)}: \mathbb{R}^V \rightarrow \mathbb{R}^U$ be the encoder of a client model $L^{(k)}$. Let $f_{c}^{(k)}: \mathbb{R}^U \rightarrow \mathbb{R}^C$ be the class classifier of $L^{(k)}$. Then, given an input sample $x$, the client model outputs $\hat{y}=f_{c}^{(k)}f_{e}^{(k)}(x)$. Similarly, the global model consists of an encoder $f_{e}^G$ and a class classifier $f_{c}^G$.  

To learn an encoder $f_{e}^{(k)}$ that disentangles the domain-invariant features from $X_S^{(k)}$, we devise the global features disentangler by introducing a domain classifier in the PS. Notably, let $f_{d}$ be the domain classifier that takes the feature representations $H$ as the input and outputs a binary variable (domain label) $q$ for each input sample $h$, which indicates whether $h$ comes from the client $k$ ($h\in H^{(k)}=f_{e}^{(k)}(X_S^{(k)}) \mbox{ if } q = 0$) or from the target domain in the PS ($h\in H^G=f_{e}^G(X_T) \mbox{ if } q = 1$). The goal of the features disentangler is to learn a neural network that distinguishes between $H^{(k)}$ and $H^G$ for different clients $k\in K$. The model learning of the features disentangler with respect to client $k$'s source domain $D_S^{(k)}$ can be formulated by the following 
\begin{equation}
J_{\mbox{disentangler}}^{(k)}=J(0, f_{d} f_{e}^{(k)}(X^{(k)}_S))+J(1, f_{d} f_{e}^{G}(X_T))
\end{equation}

\begin{equation}
\hat{f_{d}}=\underset{f_{d}}{\mbox{arg }min}J_{\mbox{disentangler}}^{(k)}(f_{d},\hat{f_{e}^{(k)}},\hat{f_{e}^{G}}),
\end{equation}

where $J$ is the negative log likelihood loss for the domain classification to identify between the representations of the source domain and the target domain.

Moreover, to learn an encoder $f_{e}^{(k)}$ that extracts domain-invariant features from client $k$'s data, $f_{e}^{(k)}$ is updated by maximizing the above classification loss $J_{\mbox{disentangler}}^{(k)}$ of the features disentangler. When the features disentangler cannot distinguish whether an input representation is from the client domain or the cloud domain, $f_{e}^{(k)}$ outputs feature vectors that are close to the ones from the target domain. Then, each client $k$ sends the feature representations ${f_{e,t}^{(k)}}(X_S^{(k)})$ to the PS every round $t$. In particular, the update of client $k$'s encoder based on the features disentangler's classification loss can be formulated by
\begin{equation}
\hat{f_{e}^{(k)}} = \underset{f_{e}^{(k)}}{\mbox{arg max}}J_{\mbox{disentangler}}^{(k)}(\hat{f_{d}},f_{e}^{(k)},\hat{f_{e}^G}).
\end{equation}


\subsubsection{Embedding Matching}
\label{sec:mmd}
The global disentangler encourages a local model to learn features that are domain-invariant. We further enhance the disentanglement of features by measuring the high-dimensional distribution difference between feature representations from a client and the target domain in the parameter server (PS). In particular, we employ the Multiple Kernel variant of Maximum Mean Discrepancy (MK-MMD) to perform embedding matching between the two distributions $H_S^{(k)}=f_{e}^{(k)}(X^{(k)}_S)$ and $H_T=f_{e}^G(X_T)$, using different Gaussian kernel $e_r, r \in \{1,2,...,R\}$ where $R$ is the number of kernels. Then, for each kernel $e_r$:

\begin{equation}
\mbox{MMD}_{e_r}^2(H_S^{(k)},H_T)=\Bigg\| \frac{1}{N^{(k)}}\sum_{i=1}^{N^{(k)}}\Phi(h_i)-\frac{1}{N^T}\sum_{j=1}^{N^T}\Phi(h_j)\Bigg\|^2_{\mathcal{H}_{e_r}}
\end{equation}

\begin{equation*}
=\frac{1}{N^{(k)}}\sum_{i=1}^{N^{(k)}}e_r(h_i,h_i')+\frac{1}{N^T}\sum_{j=1}^{N^T}e_r(h_j,h_j')-2\frac{1}{N^{(k)}} \frac{1}{N^T}\sum_{i=1}^{N^{(k)}}\sum_{j=1}^{N^T}e_r(h_i,h_j),
\end{equation*}

where $\mathcal{H}$ is the reproducing kernel Hilbert space (RKHS) and $\Phi$ is a feature map $H\rightarrow \mathcal{H}$.

Furthermore, we consider as the embedding matching loss the distance between the mean embeddings of $\Phi_{e_r}(H_S^{(k)})$ and $\Phi_{e_r}(H_T)$ with five different Gaussian kernels (a bandwidth $\mu$ of two). Then, the local model of client $k$ can be updated based on the embedding matching loss by the following
\begin{equation}
J_{\mbox{mmd}}^{(k)}=\frac{1}{5}\sum_{r=1}^{5}\mbox{MMD}_{e_r}^2(f_{e}^{(k)}(X^{(k)}_S),f_{e}^{G}(X_T))
\end{equation}	
\begin{equation}
\hat{f_{e}^{(k)}}=\underset{f_{e}^{(k)}}{\mbox{arg min}}J_{\mbox{mmd}}^{(k)}(f_{e}^{(k)},\hat{f_{e}^G}). 
\end{equation}

\subsubsection{Local Model Representation Learning}
For each round in FL, the PS sends back the computed gradients from the global feature disentangler $\frac{\partial J_{\mbox{disentangler}}^{(k)}}{\partial f_{e}^{(k)}}$ and the MK-MMD loss $J_{\mbox{mmd}}^{(k)}$ to client $k$ to update the local encoder $f_{e}^{(k)}$.
Then, for each client $k$, the local model is updated based on three different losses, i.e., the empirical loss $J^{(k)} = J(Y_S^{(k)},f_{c}^{(k)}f_{e}^{(k)}(X_S^{(k)}))$, the features disentangler loss $J_{\mbox{disentangler}}^{(k)}(\hat{f_{d}},f_{e}^{(k)},\hat{f_{e}^{G}})$, and the MK-MMD loss $J_{\mbox{mmd}}^{(k)}(f_{e}^{(k)},\hat{f_{e}^{G}}))$. To alleviate the effect of noisy representations at early stages of the training, we adopt a coefficient $\lambda_p$ that gradually changes from 0 to 1 with the learning progress of FL. Let $b$ be the batch number, $B$ be the number of total batches, $r$ be the round number, and $R$ be the number of total rounds in FL. $\lambda_p$ is defined by $\lambda_p = \frac{2}{1+exp(-\gamma \cdot p)}-1$, where $p = \frac{b+r\times B}{R\times B}$ and $\gamma$ is set as five. Notably, we devise the local model representation learning as follows
\begin{equation}
E(f_{c}^{(k)},f_{e}^{(k)}) = J^{(k)}-\lambda_p(J_{\mbox{disentangler}}^{(k)}(\hat{f_{d}},f_{e}^{(k)},\hat{f_{e}^{G}})-J_{\mbox{mmd}}^{(k)}(f_{e}^{(k)},\hat{f_{e}^{G}})) 
\end{equation}

\begin{equation}
(\hat{f_{c}^{(k)}},\hat{f_{e}^{(k)}})=\underset{f_{c}^{(k)},f_{e}^{(k)}}{\mbox{arg min}}E(f_{c}^{(k)},f_{e}^{(k)}).
\end{equation}

Then, each client $k\in K$ with the source domain $D^{(k)}_S$ performs model learning every round $t$ by the following $L^{(k)}_{t+1}=L^{(k)}_t -\eta\cdot\nabla E(f_{c}^{(k)},f_{e}^{(k)})$, where $\eta$ denotes the learning rate.

\subsubsection{Global Model Fine-Tuning Based on Federated Voting}
We propose a fine-tuning method called federated voting to update the global model every round without ground truth labels of the target domain samples. In particular, Federated voting fine-tunes the global model based on the pseudo-labels generated by the consensus from learned client local models. This strategy allows the global model to learn representations of the target domain data without ground truth labels thus improving the effectiveness of the feature distribution matching.

Let $y^{(k)}_i = \mbox{arg}\max_j f_k(x_i)_j$ represents the prediction class of client $k$'s local model $L^{(k)}$ with the input $x_i$. In FL, at each time step $t$, all client model updates $\{L^{(1)}_t-G_{t-1}, L^{(2)}_t-G_{t-1}, ..., L^{(K)}_t-G_{t-1}\}$ are uploaded to the PS. Given an unlabeled input sample $x_i$ from the target domain $D_T$, the federated voting method aims to attain the optimized classification label $y_i^*$ by the following
\begin{equation}
  y_i^* = \mbox{arg max}_{c\in \{1,2,...,C\}}\sum_{k=1}^{K}\mathbbm{1}\{y^{(k)}_i=c\}.  
  \label{eq:fv}
\end{equation}

Note that this method could result in multiple $y_i^*$ candidates that receive the same number of votes, especially when the total client number $K$ is low while the total class number $C$ is high. In such a case, we randomly select one class from the candidate pool as the label $y_i^*$ of $x_i$.

Furthermore, the fine-tuning of the global model $G_{t+1}$ is performed every round after the model aggregation with $N^T$ samples from the target domain data $X_T$ and the generated labels $Y^*=\{y_i^*\}_{i=1}^{N^T}$ using Eq. \ref{eq:fv}. We devise the global model fine-tuning as follows $G^*_{t+1} = G_{t+1}-\lambda_p\eta\cdot\nabla J(Y^*,G_{t+1}(\{x_i\}_{i=1}^{N^T}))$, where $\lambda_p$ is the coefficient to alleviate noisy voting results at early stages of FL.

\subsection{Criteria}
\label{GE}
We present two different criteria for measuring the effectiveness of a multi-source domain adaptation method, i.e., target task accuracy (TTA) and Group Effect (GE).

The performance of the global model $G_t$ at time step $t$ in the target task is measured by the target task accuracy (TTA) defined in the following
\begin{equation}
\mbox{TTA}_f(G_t) = \frac{  \sum_{ (x,y) \in D_T} \mathbbm{1}\{ \mbox{arg max}_{j}f(x;G_t)_j = y  \} }{| D_T |},
\label{eq:tta}
\end{equation}

where $|\cdot|$ denotes the size of the target domain dataset.  

Moreover, we define a novel criterion for measuring negative transfer in FL, called Group Effect (GE). In particular, GE throws light on negative transfer of FL, due to inefficient model aggregation in the PS, which has not yet been studied to our best knowledge. We aim to measure to what extent the difference in clients' training data causes diverse local model updates that eventually cancel out in the parameter space leading to negative transfer. Intuitively, FedKA matches feature distributions of separate client domains with the target domain in the parameter server, such that we can alleviate the information loss from the model aggregation in FL. Given $K$ local updates at the time step $t$, the group effect exists if $\mbox{TTA}_f(G_{t+1})<\frac{1}{K}\sum_{k\in \{1,2,...,K\}}\mbox{TTA}_f(G_{t}+\Delta^{(k)}_t)$ where $G_{t+1}$ is attained by Eq. \ref{al:fedavg}. To this end, we propose GE based on $\mbox{TTA}_f$ in Eq. \ref{eq:tta} as follows

\begin{equation}
    \mbox{GE}_t = \frac{1}{K}\sum_{k\in \{1,2,...,K\}}\mbox{TTA}_f(G_{t}+\Delta^{(k)}_t) - \mbox{TTA}_f(G_{t+1}).
    \label{eq:ge}
\end{equation}

\section{Experiments}
\label{sec:exp}
In this section, we first describe three benchmark datasets and detailed experiment settings. Next, we demonstrate the empirical evaluation results of our method using the metric of $\mbox{TTA}_f(G_t)$ in Eq.\ref{eq:tta}, followed by a discussion. Then, we compare the evaluation results based on different backbone encoder models. Furthermore, we show the effectiveness of our method in alleviating the Group Effect in Eq.\ref{eq:ge} of federated learning with both numerical results and the visualization of t-SNE features. We use PyTorch \citep{pytorch} to implement the models in this study. 

\subsection{Dataset}
We employed three domain adaptation datasets, i.e., Digit-Five and Office-Caltech10 for the image classification tasks and Amazon Review for the text classification task, respectively.

\paragraph{Digit-Five} is a collection of five most popular digit datasets, MNIST (mt) \citep{mnist} (55000 samples), MNIST-M (mm) (55000 samples), Synthetic Digits (syn) \citep{ganin} (25000 samples), SVHN (sv)(73257 samples), and USPS (up) (7438 samples). Each digit dataset includes a different style of 0-9 digit images.

\paragraph{Office-Caltech10}\citep{office} contains images of 10 categories in four domains: Caltech (C) (1123 samples), Amazon (A) (958 samples), Webcam (W) (295 samples), and DSLR (D) (157 samples). The 10 categories in the dataset consist of objects in office settings, such as keyboards, monitors, and headphones.

\paragraph{Amazon Review}\citep{Blitzer} tackles the task of identifying the sentiment of a product review (positive or negative). This dataset includes reviews from four different merchandise categories: Books (B) (2834 samples), DVDs (D) (1199 samples), Electronics (E) (1883 samples), and Kitchen \& housewares (K) (1755 samples).

\subsection{Model Architecture and Hyperparameters}
We consider different transfer learning tasks in the aforementioned datasets. Notably, we adopt each data domain in the applied dataset as a client domain. In this regard, besides the target domain of the cloud, there are four different client domains in Digit-Five and three different client domains in both Office-Caltech10 and Amazon Review, respectively. We conducted experiments using the following model architectures and hyperparameters.

\subsubsection{Image Classification Tasks}
Images in Digit-Five and Office-Caltech10 are converted to three-channel color images with a size of $28\times 28$. Then, as the backbone model, we adopt a two-layer convolutional neural network (64 and 50 channels for each layer) with batch normalization and max pooling as the encoder and two independent two-layer fully connected neural networks (100 hidden units) with batch normalization as the class classifier and the domain classifier, respectively. Moreover, to perform the local model representation learning, we apply as a learning function Adam with a learning rate of 0.0003 and a batch size of 16 based on the grid search. Every round of FL, a client performs training for one epoch using 512 random samples (32 batches) drawn from its source domain. Furthermore, to compute the features disentangler loss, 512 random samples from the target domain are applied every round. The learning of the domain classifier neural network is performed during the client model representation learning based on a gradient reversal layer.

To attain a more accurate measurement of differences in feature representation distributions using embedding matching, we apply a lager batch size of 128 with the same samples in the client model representation learning. In Digit-Five, we employ two variants of the federated voting strategy, i.e., Voting-S and Voting-L, using 512 and 2048 random target domain samples, respectively. Moreover, since Office-Caltech10 is a relatively small dataset, we use all available samples in the target domain for federated voting. The learning hyperparameters of the global model are the same with the local models.

\subsubsection{Text Sentiment Classification Task}
To process text data of product reviews, we apply the pretrained Bidirectional Encoder Representations from Transformers (BERT) \citep{bert} to convert the reviews into 768-dimensional embeddings. We set the longest embedding length to 256, cutting the excess and padding with zero vectors. Moreover, we apply a two-layer fully connected neural network (500 hidden units) with batch normalization as the encoder using the flatten embeddings as the input. The features disentangler and class classifier share the same architectures as those in the image classification tasks, but with different input and output shapes (binary classification). We apply Adam with a learning rate of 0.0003, a batch size of 16, and 128 random training samples every round. A lower training sample number is because Amazon Review has much fewer samples compared to Digit-Five. Similarly, we employ 128 random target domain samples every round for federated voting. In addition, in embedding matching, we apply a batch size of 16.

\subsection{Ablation Study}
To understand different components' effectiveness in FedKA, we performed an ablation study by evaluating $max_{G_t} \mbox{TTA}_f(G_t)$ during 200 rounds of FL. We considered different combinations of the three building blocks of FedKA and evaluated their effectiveness in different datasets. As a comparison model, the FedAvg method applies the averaged local updates to update the global model. Moreover, the f-DANN method extends Domain Adversarial Neural Network (DANN) to the specific task of federated learning, where each client has an individual DANN model for training. Similarly, for the f-DAN method, we adapted Deep Adaptation Network (DAN) to the specific task of federated learning. We discuss the evaluation results of the ablation study in the following.

Table \ref{tab:abldigit} demonstrates the evaluation results in Digit-Five. Though f-DANN improved the model performance in the tasks of $mt,sv,sy,up\rightarrow mm$ and $mt,mm,sv,up\rightarrow sy$, f-DANN resulted in a decreased $max_{G_t} \mbox{TTA}_f(G_t)$ in the other three tasks. Furthermore, for the two variants of federated voting, the results suggest that federated voting can improve the model performance, especially, combined with the global feature disentangler and embedding matching. In the experiment on the Digit-Five dataset, FedKA achieves the best accuracy improving model performance by 6.7\% on average.


Table \ref{tab:office} demonstrates the evaluation results in Office-Caltech10. The proposed method also outperforms the other comparison models. In addition, the effectiveness of federated voting appears to smaller compared to the case of Digit-Five. This is due to the less available target data in Office-Caltech10 for fine-tuning the global model.

Table \ref{tab:ablnlp} demonstrates the evaluation results in Amazon Review. FedKA improved the global model's performance in all target tasks outperforming the other approaches, by leveraging the global feature disentangler, embedding matching, and federated voting.


\begin{table*}[!t]
    \centering
    \scriptsize
    \renewcommand{\arraystretch}{1.15}
    \caption{$max_{G_t} \mbox{TTA}_f(G_t)$ ($\%$) on the Digit-Five dataset based on different methods. The highest reported accuracy under each task is in bold.}
    \label{tab:abldigit}
    \begin{tabular}{lcccccc}
    \hline
        Models/Tasks & $\rightarrow$mt & $\rightarrow$mm & $\rightarrow$up & $\rightarrow$sv & $\rightarrow$sy & Avg\\ \hline
        FedAvg & $\underbar{93.5}\pm$0.15 & 62.5$\pm$0.72 & 90.2$\pm$0.37 & 12.6$\pm$0.31& 40.9$\pm$0.50& 59.9\\
        f-DANN & 89.7$ \pm$0.23 & 70.4$ \pm$0.69 & 88.0$ \pm$0.23 & 11.9$ \pm$0.50 & 43.8$ \pm$1.04 & 60.8\\ 
        f-DAN& $\underbar{93.5} \pm$0.26& 62.1$ \pm$0.45& 90.2$ \pm$0.13 & 12.1$ \pm$0.56 & 41.5$ \pm$0.76& 59.9\\ 
        \hline
        Voting-S & $\textbf{93.7} \pm$0.18 & 63.4$ \pm$0.28$ $&$\textbf{92.6} \pm$0.25 & 14.2$ \pm$0.99$ $&45.3$ \pm$0.34&61.8\\
        Voting-L & $\underbar{93.5} \pm$0.18 &64.8$ \pm$1.01 & $\underbar{92.3} \pm$0.21 & 14.3$ \pm$0.42$ $& 45.6$ \pm$0.57 & 62.1\\ 
        Disentangler + Voting-S & 91.8$ \pm$0.20 &71.2$ \pm$0.40 & 91.0$ \pm$0.58 & 14.4$ \pm$1.09 & 48.7$ \pm$1.19 & 63.4\\ 
        Disentangler + Voting-L & 92.1$ \pm$0.16 & $\underbar{71.8} \pm$0.48 & 90.9$ \pm$0.36 & $\underbar{15.1} \pm$0.91 & $\underbar{49.1} \pm$1.03 & $\underbar{63.8}$\\ 
        Disentangler + MK-MMD & 90.0$ \pm$0.49 & 70.4$ \pm$0.86 & 87.5$ \pm$0.25 & 12.2$ \pm$0.70& 44.3$ \pm$1.18$ $& 60.9\\ 
        FedKA-S & 91.8$ \pm$0.19 & $\underbar{72.5} \pm$0.91 & 90.6$ \pm$0.14 & $\textbf{15.2} \pm$0.46 & $\underbar{48.9} \pm$0.48 & $\underbar{63.8}$\\ 
        FedKA-L 	& 92.0$ \pm$0.26	& $\textbf{72.6} \pm$1.03 & $\underbar{91.1} \pm$0.24$ $& $\underbar{14.8} \pm$0.41 & $\textbf{49.2} \pm$0.78 & $\textbf{63.9}$\\ \hline
    \end{tabular}
    \end{table*}
    
\begin{table*}[!t]
   \centering
    \scriptsize
   \renewcommand{\arraystretch}{1.15}
    \caption{$max_{G_t} \mbox{TTA}_f(G_t)$ ($\%$) on the Office-Caltech10 dataset based on different methods. The highest reported accuracy under each task is in bold.}
    \label{tab:office}
    \begin{tabular}{lcccccc}
    \hline
        Models/Tasks & C,D,W$\rightarrow$A & A,D,W$\rightarrow$C & C,A,W$\rightarrow$D & C,D,A$\rightarrow$W & Avg \\ \hline
        FedAvg & $\underbar{52.9}$ $\pm$0.56 & $\textbf{37.5}$ $\pm$0.50 & 28.7$\pm$1.80 & 22.4$\pm$1.38 & 35.4 \\
        f-DANN & $\underbar{52.8}$ $\pm$0.40 & $\underbar{37.3}$ $\pm$0.84 & 28.8$\pm$2.07 & 23.3$\pm$0.51 & 35.5 \\
        f-DAN & 52.7$\pm$0.64 & 36.8$\pm$0.49 & 28.4$\pm$1.43 & 22.9$\pm$0.76 & 35.2 \\
        \hline
        Voting & $\textbf{53.3}$ $\pm$0.80 & $\underbar{37.3}$ $\pm$0.58 & 27.8$\pm$2.37 & 23.3$\pm$1.92 & 35.4 \\ 
        Disentangler + Voting & 52.5$\pm$0.65 & $\underbar{37.3}$ $\pm$0.84 & $\underbar{29.9}$ $\pm$2.70 & $\underbar{23.4}$ $\pm$1.72 & $\underbar{35.8}$ \\
        Disentangler + MK-MMD & 52.7$\pm$0.41 & 36.4$\pm$0.93 & $\textbf{31.1}$ $\pm$1.91 & $\textbf{24.3}$ $\pm$1.69 & $\textbf{36.1}$ \\ 
        FedKA & $\underbar{52.8}$ $\pm$0.57 & 37.2$\pm$0.29 & $\underbar{29.3}$ $\pm$1.51 & $\underbar{23.7}$ $\pm$1.15 & $\underbar{35.8}$ \\ \hline
    \end{tabular}
\end{table*}

    \begin{table*}[!t]
    \centering
    \scriptsize
    \renewcommand{\arraystretch}{1.15}
    \caption{$max_{G_t} \mbox{TTA}_f(G_t)$ ($\%$) on the Amazon Review dataset based on different methods. The highest reported accuracy under each task is in bold.}
    \label{tab:ablnlp}
    \begin{tabular}{lcccccc}
    \hline
        Models/Tasks & D,E,K$\rightarrow$B & B,E,K$\rightarrow$D & B,D,K$\rightarrow$E & B,D,E$\rightarrow$K & Avg\\ \hline
        FedAvg & 62.6$\pm$0.58 & 75.1$\pm$0.53 & 78.0$\pm$0.39 & 80.3$\pm$0.34 & 74\\ 
        f-DANN & $\underbar{62.7}\pm$0.35 & $\underbar{75.3}\pm$0.34 & $\underbar{78.7}\pm$0.29 & $\underbar{80.4}\pm$0.21 & $\underbar{74.3}$\\
        f-DAN & 62.3$\pm$0.55 & 73.8$\pm$0.29 & 77.8$\pm$0.29 & 80.1$\pm$0.38 & 73.5\\
        \hline
        Voting & 62.1$\pm$0.20 & 74.6$\pm$0.58 & 77.8$\pm$0.70 & 79.6$\pm$0.33 & 73.5\\ 
        Disentangler + Voting & $\underbar{62.7}\pm$0.37 & 75.1$\pm$0.60 & 78.4$\pm$0.29 & $\textbf{80.8}\pm$0.52 & $\underbar{74.3}$\\ 
        Disentangler + MK-MMD & $\textbf{62.9}\pm$0.32 & $\underbar{75.3}\pm$0.33 & $\underbar{78.5}\pm$0.23 & 80.2$\pm$0.12 & 74.2\\ 
        FedKA & $\underbar{62.8}\pm$0.23 & $\textbf{75.8}\pm$0.52 & $\textbf{78.8}\pm$0.65 & $\underbar{80.7}\pm$0.28 & $\textbf{74.5}$\\ \hline
    \end{tabular}
\end{table*}

\subsection{Effectiveness of Model Architecture Complexity}
\label{sec:extractor}
\begin{figure}[!t]
    \centering
    \includegraphics[width=0.6\linewidth]{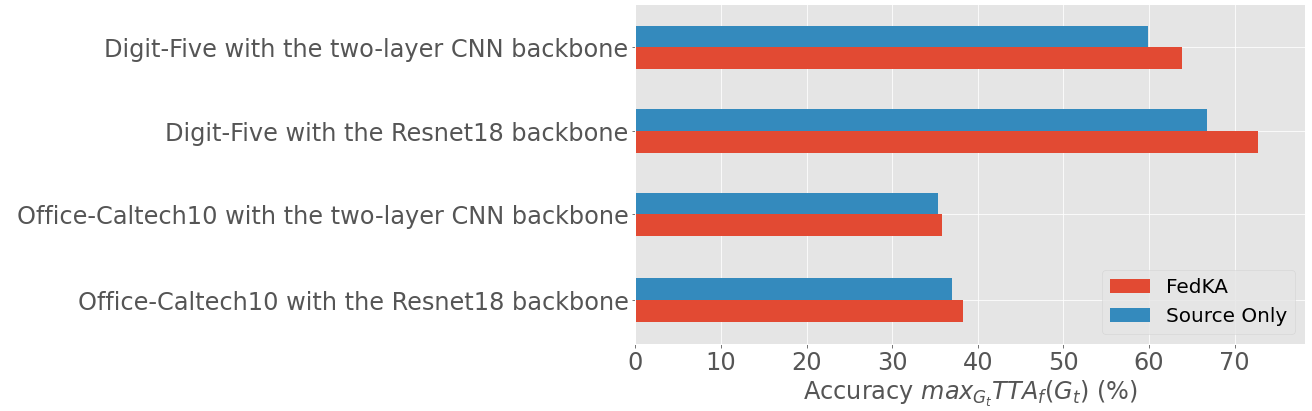}
    \caption{Model performance comparison between the Source Only (FedAvg) method and FedKA using different backbone encoder models. The result shows that FedKA can better benefit the improvement in global model performance with a more complex encoder model.}
    \label{fig:comparision}
\end{figure}

To further study the effectiveness of Federated Knowledge Alignment (FedKA) when applying different encoder models, we employed Resnet18 \citep{resnet} without pretraining to perform feature extraction. Based on the same hyperparameter setting, we evaluate the model performance in Digit-Five (Table \ref{tab:resnetdigit}) and Office-Caltech10 (Table \ref{tab:resnetoffice}). As a result, FedKA achieved performance gains of 8.8\% and 3.5\% in Digit-Five and Office-Caltech10 with the Resnet18 backbone, respectively.
Moreover, as shown in Figure \ref{fig:comparision}, a more complex model could contribute to a larger gain in the global model performance improvement.

\begin{table*}[!t]
    \centering
    \scriptsize
    \renewcommand{\arraystretch}{1.15}
    \caption{$max_{G_t} \mbox{TTA}_f(G_t)$ ($\%$) on the Digit-Five dataset using Resnet18 as the backbone. The highest reported accuracy under each task is in bold.}
    \label{tab:resnetdigit}
    \begin{tabular}{lcccccc}
    \hline
        Models/Tasks & $\rightarrow$mt & $\rightarrow$mm & $\rightarrow$up & $\rightarrow$sv & $\rightarrow$sy & Avg\\ \hline
        FedAvg & $\textbf{97.9}\pm$0.07 & 71.3$\pm$0.79 & $\textbf{96.9}\pm$0.05 & 11.9$\pm$0.62& 55.8$\pm$1.60& 66.8\\ 
        f-DANN & $\underbar{97.5} \pm$0.07 & $\underbar{77.1} \pm$0.29 & 96.8$ \pm$0.38 & 12.1$ \pm$1.01 & $\underbar{79.5} \pm$0.37 & 72.6\\ 
        f-DAN & $\textbf{97.9} \pm$0.09& 71.7$ \pm$1.22& 96.7$ \pm$0.18 & 11.3$ \pm$0.68 & 55.5$ \pm$1.00& 66.5\\ 
        \hline
        Voting & 96.5$ \pm$0.20 & 72.1$ \pm$1.24$ $&$\textbf{96.9} \pm$0.17 & $\textbf{14.0} \pm$0.74$ $&61.1$ \pm$0.28 & 68.1\\
        Disentangler + Voting & 96.5$ \pm$0.21 &76.5$ \pm$0.53 & 96.8$ \pm$0.42 & $\underbar{13.7} \pm$0.45 & 79.4$ \pm$0.61 & $\underbar{72.6}$\\ 
        Disentangler + MK-MMD & $\underbar{97.5} \pm$0.03 & $\underbar{76.7} \pm$0.69 & $\textbf{96.9} \pm$0.15 & 11.0$ \pm$0.53& $\textbf{80.1} \pm$0.49$ $& 72.4\\ 
        FedKA & 96.4$ \pm$0.23 & $\textbf{77.3} \pm$1.01 & 96.6$ \pm$0.38 & $\underbar{13.8} \pm$0.81 & $\underbar{79.5} \pm$0.68 & $\textbf{72.7}$\\
        \hline
    \end{tabular}
    \end{table*}

\begin{table*}[!t]
    \centering
    \scriptsize
    \renewcommand{\arraystretch}{1.15}
    \caption{$max_{G_t} \mbox{TTA}_f(G_t)$ ($\%$) on the Office-Caltech10 dataset using Resnet18 as the backbone. The highest reported accuracy under each task is in bold.}
    \label{tab:resnetoffice}
    \begin{tabular}{lcccccc}
    \hline
        Models/Tasks & C,D,W$\rightarrow$A & A,D,W$\rightarrow$C & C,A,W$\rightarrow$D & C,D,A$\rightarrow$W & Avg \\ \hline
        FedAvg & 56.4 $\pm$1.23 & $\underbar{40.2}$ $\pm$0.69 & 28.7$\pm$1.21 & 22.7$\pm$1.85 & 37.0 \\
        f-DANN & 58.3 $\pm$1.53 & 40.0 $\pm$1.50 & $\underbar{30.7}$ $\pm$3.59 & 22.3$\pm$1.29 & $37.8$ \\
        f-DAN  & 56.7$\pm$0.71 & 38.7$\pm$0.75 & 30.2$\pm$1.64 & $\underbar{23.9}$ $\pm$1.70 & 37.4 \\ 
        \hline
        Voting & 56.5 $\pm$1.88 & $\underbar{40.2}$ $\pm$0.58 & 29.8$\pm$1.45 & $\textbf{24.1}$ $\pm$0.69 & 37.7 \\ 
        Disentangler + Voting & $\textbf{61.4}$ $\pm$2.51 & $\textbf{40.4}$ $\pm$1.01 & $\underbar{31.5}$ $\pm$3.11 & $\underbar{23.9}$ $\pm$1.89 & $\textbf{39.3}$ \\
        Disentangler + MK-MMD & $\underbar{59.5}$ $\pm$0.41 & 37.8$\pm$0.93 & $\textbf{32.2}$ $\pm$3.21 & 22.3 $\pm$1.00 & $\underbar{38.0}$ \\ 
        FedKA & $\underbar{59.9}$ $\pm$1.44 & 39.7$\pm$0.81 & 30.2 $\pm$1.71 & 23.4 $\pm$1.45 & \underbar{38.3} \\ \hline
    \end{tabular}
\end{table*}

\subsection{Effectiveness in Alleviating the Group Effect of Federated Learning}
\label{exp:GE}
\begin{figure}[!t]
    \centering
  \subfloat[mt,sv,sy,up$\rightarrow$mm\label{1a}]{%
       \includegraphics[width=0.35\linewidth]{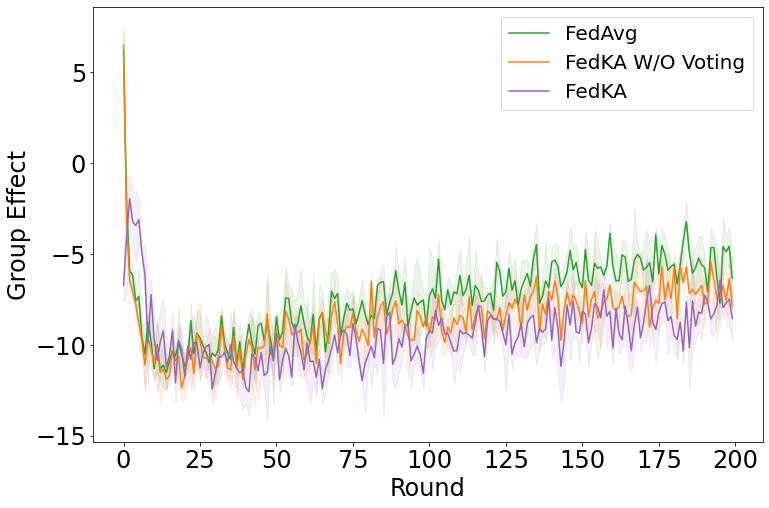}}
     \qquad
  \subfloat[mt,mm,sv,up$\rightarrow$sy\label{1b}]{%
    \includegraphics[width=0.36\linewidth]{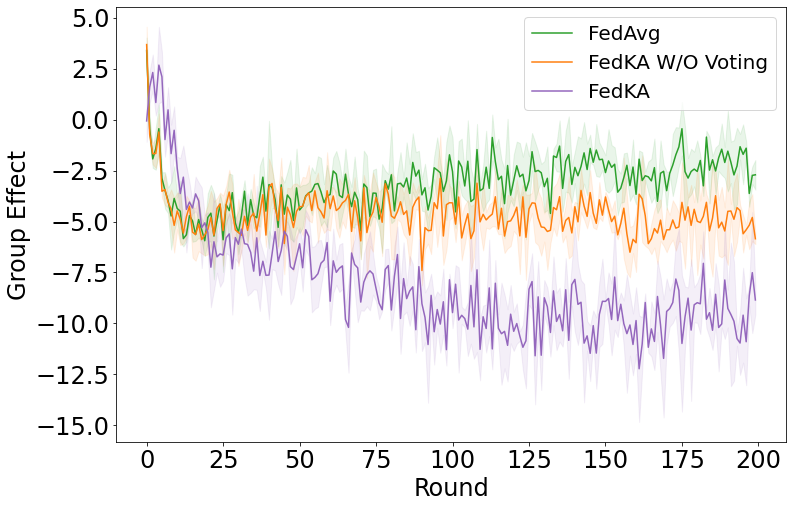}}
  \caption{Group Effect $\mbox{GE}_t$ during the 200 rounds of FL. Lower is better. In Figure (a), though we observed a bounce-back behavior of GE after 20 rounds, the negative transfer in each round was decreased by FedKA. Moreover, in Figure (b), FedKA kept reducing GE even after FedAvg started to bounce back, successfully alleviating the negative transfer of model aggregation by 1 time without voting and 4 times with voting, respectively. Compared to the usual model aggregation of FedAvg, FedKA showed great performance in alleviating Group Effect by the global model fine-tuning with federated voting.}
  \label{groupeffect}
\end{figure}

To understand the effectiveness of FedKA in alleviating the negative transfer in Federated Learning (FL), we evaluated Group Effect $\mbox{GE}_t$ in Eq. \ref{eq:ge} during the 200 rounds of FL with the Digit-Five dataset (Figure \ref{groupeffect}). The GE value represents the amount of negative transfer occurring in FL during model aggregation, where a higher GE value reflects more information loss from the aggregation and a negative value represents a performance gain via the aggregation. In particular, as shown in the graphs, the learning progress had high GE values at the early stages, implicating that the model aggregation results in information loss and degraded model performance. As learning progresses, the GE values keep decreasing implicating the gradual convergence of client models towards the target domain distribution. The results show that FedKA can greatly alleviate the negative transfer in the model aggregation of FL increasing the global model's performance in unseen tasks.

\begin{figure*}[!t]
    \centering
    \includegraphics[width=0.7\linewidth]{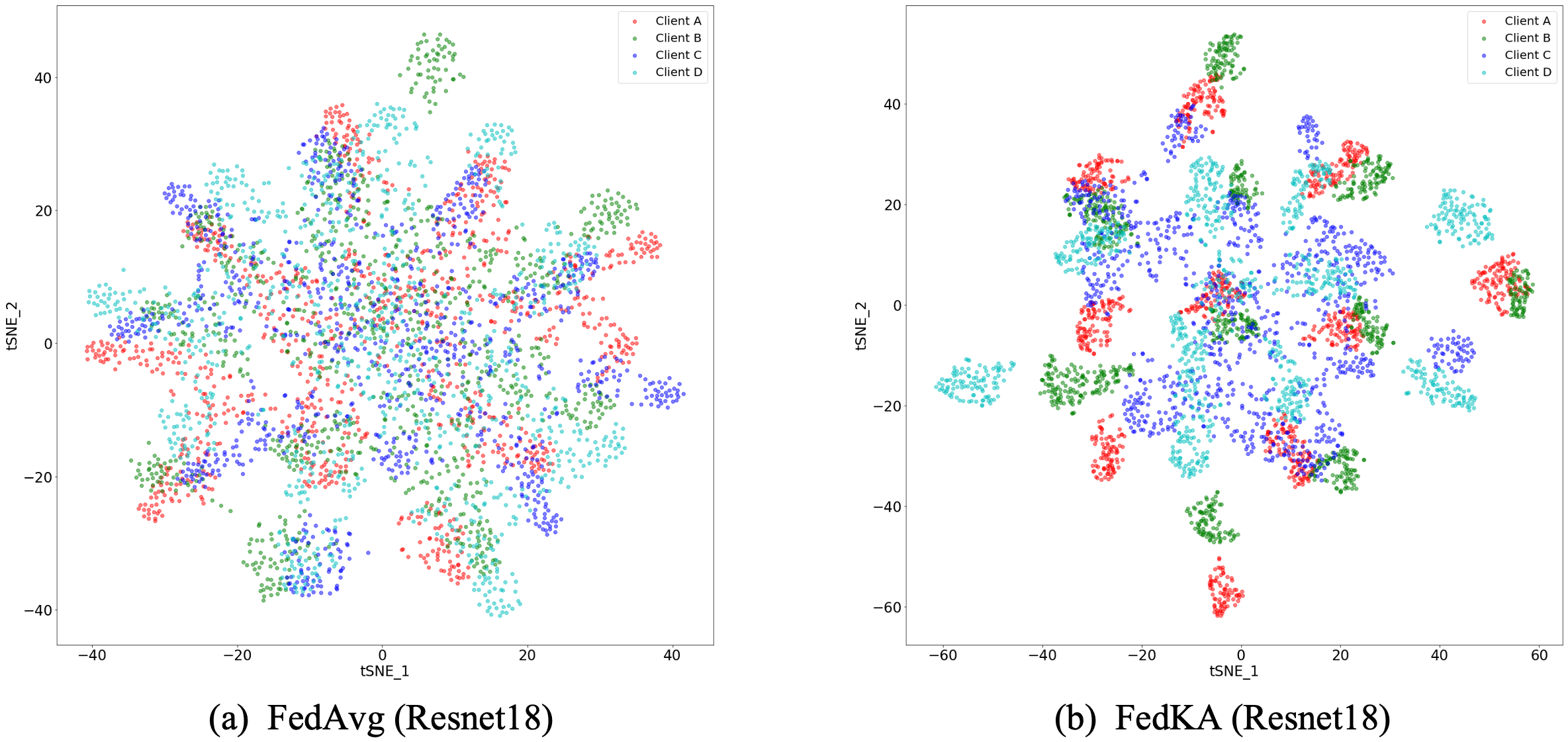}
    \caption{T-SNE visualization of different client domain feature distributions in Digit-Five when applying Resnet18 as the encoder backbone.}
    \label{fig:tsne-1}
\end{figure*}

To further verify the effectiveness of FedKA in alleviating negative transfer, we employ t-SNE \citep{tsne} to visualize the extracted feature distributions from different client data domains based on the learned global model (Figure \ref{fig:tsne-1}). Apparently, the global model based on FedKA learns better representations for the classification tasks. 

\section{Conclusion}
Federated Learning (FL) has been adopted in various walks of life to facilitate machine learning on distributed client data. Nevertheless, the data discrepancy between clients usually hinders the effectiveness of model transfer in FL. Traditional domain adaptation methods usually require prior knowledge of different domain data and cannot benefit FL under the constraint of data confidentiality. In this work, we proposed Federated Knowledge Alignment (FedKA) to allow domain feature matching in the global workspace. FedKA improves the transferability of learned domain knowledge alleviating negative transfer in FL. The extensive experiments showed that FedKA could improve the global model's generality to unseen image and text classification tasks. In future work, we aim to employ self-supervised learning methods such as contrastive learning \citep{ssl,barlow,clip} to further improve the model's performance. Moreover, we will also consider the security of the proposed framework encountered with adversarial attacks such as information stealing \citep{Huang21, Yin21} in our future study.   

\section*{Acknowledgement}
This work was partially supported by JSPS KAKENHI Grant Number JP22J12681 and JP22H03572. The authors would also like to thank United Nations University for funding this research and the anonymous reviewers for their constructive comments and suggestions.

\bibliographystyle{unsrtnat}
\bibliography{acml22}
\end{document}